%% file: main_paper.tex
\documentclass[nohyperref]{article}

\usepackage{microtype}
\usepackage{booktabs} 
\usepackage{xspace}

\usepackage{hyperref}

\usepackage[accepted]{icml2022}

\usepackage{hyperref}
\usepackage{url}
\usepackage{xspace}
\usepackage{dsfont}
\usepackage{amsthm}
\usepackage{enumitem}
\usepackage{stmaryrd}
\usepackage{booktabs}
\usepackage{amssymb}
\usepackage{mathtools}
\usepackage{amsfonts}
\usepackage{pifont}
\usepackage{xfrac}
\usepackage{nicefrac}

\usepackage{graphicx}
\usepackage{caption}
\usepackage{subcaption}

\usepackage[capitalize,noabbrev]{cleveref}

\input{sections/math_commands}

\input{sections/notations}

\theoremstyle{plain}

\theoremstyle{definition}

\theoremstyle{remark}

\usepackage[textsize=tiny]{todonotes}

\icmltitlerunning{A Framework for Learning to Request Rich and Contextually Useful Information from Humans}

\begin{document}

\twocolumn[
\icmltitle{A Framework for Learning to \\ Request Rich and Contextually Useful Information from Humans}



\icmlsetsymbol{equal}{*}

\begin{icmlauthorlist}
\icmlauthor{Khanh Nguyen}{umd}
\icmlauthor{Yonatan Bisk}{cmu}
\icmlauthor{Hal Daum\'e III}{umd,msr}
\end{icmlauthorlist}

\icmlaffiliation{umd}{University of Maryland, College Park}
\icmlaffiliation{cmu}{Carnegie Mellon University}
\icmlaffiliation{msr}{Microsoft Research}

\icmlcorrespondingauthor{Khanh Nguyen}{kxnguyen@umd.edu}

\icmlkeywords{Machine Learning, Human-AI interaction, ICML}

\vskip 0.3in
]



\printAffiliationsAndNotice{}  

\begin{abstract}
\input{sections/abstract}
\end{abstract}

\setlist[itemize]{noitemsep,nolistsep}
\setlist[enumerate]{label=(\alph*),noitemsep,nolistsep}

\input{sections/intro}
\input{sections/prelim}

\input{sections/problem}
\input{sections/framework}
\input{sections/exp_setup}

\input{sections/results}
\input{sections/related}
\input{sections/conclusion}

\bibliography{journal_full,main_paper}
\bibliographystyle{icml2022}

\newpage
\appendix
\onecolumn
\input{sections/appendix_content}

\end{document}

%% file: sections/math_commands.tex
\usepackage{amsmath,amsfonts,bm}

\def\eqref#1{equation~\ref{#1}}

\def\floor#1{\lfloor #1 \rfloor}
\def\1{\bm{1}}

\DeclareMathAlphabet{\mathsfit}{\encodingdefault}{\sfdefault}{m}{sl}
\SetMathAlphabet{\mathsfit}{bold}{\encodingdefault}{\sfdefault}{bx}{n}

\def\gA{{\mathcal{A}}}
\def\gB{{\mathcal{B}}}

\def\gD{{\mathcal{D}}}

\def\gG{{\mathcal{G}}}

\def\gS{{\mathcal{S}}}
\def\gT{{\mathcal{T}}}

\def\gY{{\mathcal{Y}}}

\DeclareMathOperator*{\argmax}{arg\,max}

%% file: sections/notations.tex
\renewcommand{\b}[1]{\bar{#1}}

\newcommand{\cur}{\textsc{Cur}\xspace}
\newcommand{\goal}{\textsc{Goal}\xspace}
\newcommand{\sub}{\textsc{Sub}\xspace}
\newcommand{\doa}{\textsc{Do}\xspace}
\newcommand{\done}{\textsc{Done}\xspace}

\newcommand{\unseenstr}{\textsc{UnseenStr}\xspace}
\newcommand{\unseenobj}{\textsc{UnseenObj}\xspace}
\newcommand{\unseenenv}{\textsc{UnseenEnv}\xspace}
\newcommand{\ti}[1]{{\tiny\textcolor{black!50}{(#1)}}}

\newcommand{\oppol}{\hat{\pi}}
\newcommand{\inpol}{\psi_{\theta}}
\newcommand{\vfunc}{V_{\phi}}

\newcommand{\han}{\textsc{Han}\xspace}
\newcommand{\framework}{\textsc{Hari}\xspace}

\renewcommand{\sectionautorefname}{\S\kern-0.2em}
\renewcommand{\subsectionautorefname}{\S\kern-0.2em}

\newcommand{\intpol}{\psi_{\theta}}
\newcommand{\intpoltxt}{intention policy\xspace}
\newcommand{\intenvtxt}{intention environment\xspace}
\newcommand{\exepol}{\hat{\pi}}
\newcommand{\exepoltxt}{execution policy\xspace}
\newcommand{\exeenvtxt}{execution environment\xspace}

%% file: sections/abstract.tex
When deployed, AI agents will encounter problems that are beyond their autonomous problem-solving capabilities. 
Leveraging human assistance can help agents overcome their inherent limitations and robustly cope with unfamiliar situations. 
We present a general interactive framework that enables an agent to request and interpret rich, contextually useful information from an assistant that has knowledge about the task and the environment. 
We demonstrate the practicality of our framework on a simulated human-assisted navigation problem.
Aided with an assistance-requesting policy learned by our method, a navigation agent achieves up to a 7$\times$ improvement in success rate on tasks that take place in previously unseen environments, compared to fully autonomous behavior. 
We show that the agent can take advantage of different types of information depending on the context, and
analyze the benefits and challenges of learning the assistance-requesting policy when the assistant can recursively decompose tasks into subtasks.\looseness=-1

%% file: sections/intro.tex
\section{Introduction}

\begin{figure*}[t!]
    \centering
    \includegraphics[width=1.05\linewidth]{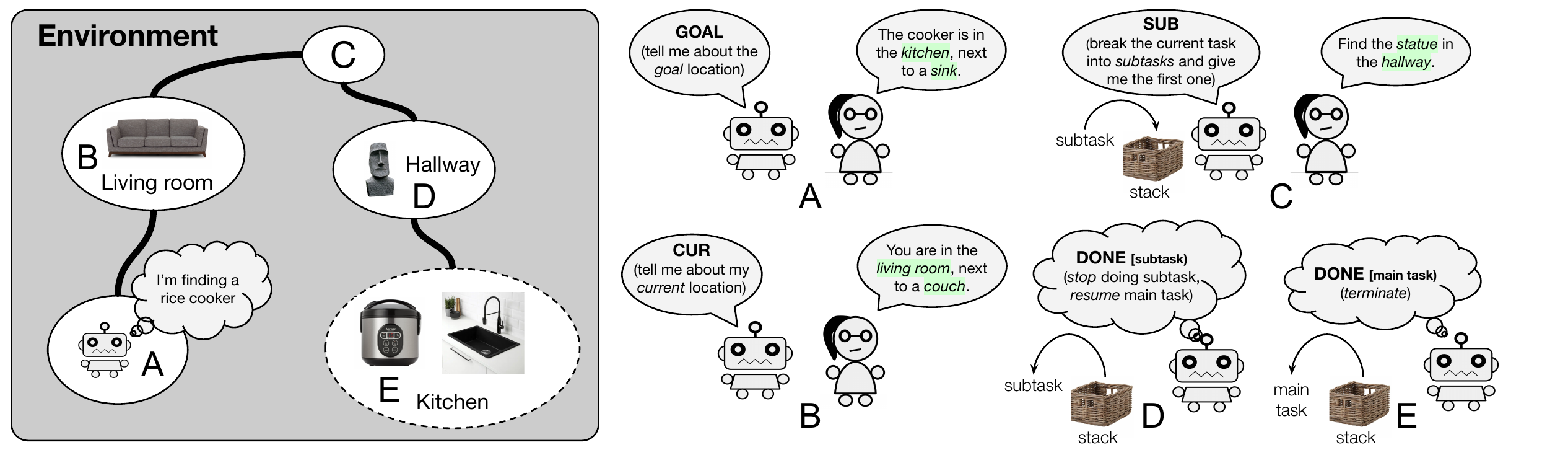}
    \caption{An illustration of the \framework framework in a navigation task. The agent can only observe part of an environment and is asked to find a rice cooker. An assistant communicates with the agent and can provide information about the environment and the task. Initially (A) the agent may request information about the goal, but may not know where it currently is. 
    For example, at location B, due to limited perception, it does not recognize that it is next to a couch in a living room. It can obtain such information from the assistant. If the current task becomes too difficult (like at location C), the agent can ask the assistant for a simpler subtask that helps it make progress toward the goal. 
    When the agent receives a subtask, it pushes the subtask to the top of a goal stack, and when it decides to stop executing a subtask, it pops the subtask from the stack (e.g., at location D). At location E, the agent empties the stack and terminates its execution. }
    \label{fig:illustration}
\end{figure*}

Machine learning has largely focused on creating agents that can solve problems on their own. 
Despite much progress, these autonomous agents struggle to operate in unfamiliar environments and fulfill new requirements from users  \citep{goodfellow2014explaining,jia-liang-2017-adversarial,eykholt2018robust,qi2020reverie,shridhar2020alfred}.
As autonomous agents are built to perform tasks by themselves, they are typically endowed with limited capabilities of communicating with humans during task execution.
This design significantly constrains the performance of these agents, as human bystanders or supervisors can provide useful information that improves the agents' decisions. 
Uncommunicative autonomous agents are also highly unsafe.
They do not consult humans on difficult decisions and can potentially commit catastrophic mistakes without warnings. 

The reliability of autonomous agents can be dramatically enhanced if they are equipped with communication skills for leveraging knowledge of humans in the same environment \citep{rosenthal2010effective,Tellex-RSS-14,nguyen2019vnla,nguyen2019hanna,thomason2020vision}. 
In many cases, a task that is initially out of reach of an agent can be elaborated or re-formulated into a task that the agent is familiar with.
Through effective communication, the agent can help a human revise the task specification into a form that it can easily interpret and accomplish.
As a motivating example, consider a home-assistant robot assigned a task of finding a newly bought \textit{rice cooker} in a house, which it has never heard of and therefore does not know how to proceed. 
The robot, however, knows how to navigate to familiar landmarks in the house. 
In this case, the robot can ask a human it encounters a question like ``\textit{where is the rice cooker?}'', expecting an instruction like ``\textit{find it in the kitchen, near the sink}'', which it may know how to execute.
By communicating with the human, the robot can convert the initially impossible task into a more feasible one. 
On the other hand, by giving the right information to the agent, a human can lift its task performance without needing to collect extra data and acquire expertise to upgrade its model. \looseness=-1

We present \framework: \textbf{H}uman-\textbf{A}ssisted \textbf{R}einforced \textbf{I}nteraction, a general, POMDP-based framework that allows agents to effectively request and interpret information from humans.
We identify and provide solutions to two fundamental problems: the \textit{speaker} problem and the \textit{listener} problem.
The speaker problem concerns teaching an agent to elicit information from humans that can improve its decision making. 
Inspired the intention-based theory of human communication \citep{sperber1986relevance, tomasello2005understanding, scott2014speaking}, we equip a learning agent with a set of general information-seeking intentions that are helpful in solving any POMDP problem.
At every time step, the agent can express to a human an intention of requesting additional information about (i) its current state, (ii) the goal state which it needs to reach, or (iii) a new subgoal state which, if reached, helps it make progress on the main goal. 
The agent learns to decide what information to request through reinforcement learning, by learning how much each type of information enhances its performance in a given situation.
The agent's request decisions are thus grounded in its own perception of its (un)certainties about the current situation and are optimized to be maximally contextually useful to it. 

Our approach contrasts with methods to train agents to ask questions by mimicking those asked by humans \citep{labutov2015deep,mostafazadeh-etal-2016-generating,de2017guesswhat,rao-daume-iii-2018-learning,thomason2020vision,liu2020asking}. 
While effective in some situations, this approach has a potential significant drawback: questions asked by humans may not always be contextually useful for agents. For example, humans are experts in recognizing objects, thus rarely ask questions like ``\textit{what objects are next to me}?''
Meanwhile, such questions may be helpful for robot localization, as robots may have imperfect visual perception in unfamiliar environments. 
Rather than focusing on naturalness, we aim to endow an agent with the cognitive capability of 
determining which type of information would be contextually useful to itself.
Such capability requires an understanding of the casual effects of the agent's decisions on its future performance, which is exactly the knowledge acquired through reinforcement learning. 

In addition to being able to \textit{ask} for useful information, the agent must be able to incorporate the information it receives into its decision-making process: the \textit{listener} problem. 
Humans can offer various types of assistance and may express them using diverse media (e.g., language, gesture, image). 
\framework implements a flexible communication protocol that is defined by the agent's task-solving policy:
information coming from the human is given as state descriptions that the policy can take as input and map into actions. 
Hence, the space of information that the human can convey is as rich as the input space of the policy.
Our design takes advantage of the flexibility of constructing an agent policy: (i) the policy can be further trained to interpret new information from humans, and (ii) it can leverage modern neural network architectures to be able to encode diverse forms of information. 
This contrasts with existing approaches based on reinforcement learning (RL) or imitation learning (IL) \citep{knox2009interactively,judah2010reinforcement,torrey2013teaching,judah2014active}, in which feedback to the agent is limited to scalar feedback of rewards (in RL) or actions (in IL).\looseness=-1

To evaluate our framework, we simulate a human-assisted navigation problem where an agent can request information about the environment from a human. 
Our agent learns an \textit{intention policy} to decide at each step whether it wants to request additional information, and, if so, which type of information it wants to obtain.
On tasks in previously unseen environments, the ability to ask for help improves the agent's success rate by 7$\times$ compared to performing tasks on its own. 
This human-assisted agent even outperforms an agent that has perfect perception and goal descriptions in unseen environments, thanks to the ability to simplify difficult tasks via requesting subgoals.
We release code and data for the experiments at \url{https://github.com/khanhptnk/hari}.
\looseness=-1

%% file: sections/prelim.tex
\section{Preliminaries}
\label{sec:prelim}

We consider an environment defined by a partially observed Markov decision process (POMDP) $E = (\gS, \gA, T, c, \gD, \rho)$ with state space $\gS$, action space $\gA$, transition function $T: \gS \times \gA \rightarrow \Delta(\gS)$, cost function $c: (\gS \times \gS) \times \gA \rightarrow \mathbb{R}$, description space $\gD$, and description function $\rho: \gS \rightarrow \Delta(\gD)$, where
 $\Delta(\gY)$ denotes the set of probability distributions over $\gY$.\footnote{We say ``description'' in lieu of ``observation'' to emphasize (i) the description can come in varied modalities (e.g., image or text), and (ii) it can be obtained via perception as well as communication. }
We refer to this as the \textit{\exeenvtxt}, where the agent performs tasks---e.g., a navigation environment.

A (goal-reaching) \textit{task} is a tuple $(s_1, g_1, d^g_1)$ where $s_1$ is the start state, $g_1$ is the goal state, and $d^g_1$ is a limited description of $g_1$.\footnote{Our notion of ``goal state'' refers not only to physical goals (e.g. a location in a house),
but also to abstract states (e.g. a level of house cleanliness).}
Initially, a task $(s_1, g_1, d^g_1)$ is sampled from a distribution $\mathfrak{T}$.
An agent starts in $s_1$ and is given the goal description $d^g_1$. 
It must reach the goal state $g_1$ within $H$ steps. 
Let $g_t$ and $d^g_t$ be the goal state and goal description executed at time step $t$, respectively.
In a standard POMDP, $g_t = g_1$ and $d^g_t = d^g_1$ for $1 \leq t \leq H$. 
Later, we will enable the agent to set new goals via communication with humans.

At time $t$, the agent does not know its state $s_t$ but only receives a state description $d^s_t \sim \rho(s_t)$.
Given $d^s_t$ and $d^g_t$, the agent makes a decision $a_t \in \gA$, transitions to the next state $s_{t + 1} \sim T(s_t, a_t)$, and incurs an \textit{operation cost} $c_t \triangleq c(s_t, g_t, a_t)$.
When the agent takes the $a_{\textrm{done}}$ action to terminate its execution, it receives a \textit{task error} $c(s_t, g_t, a_{\textrm{done}})$ that indicates how far it is from actually completing the current task.
The agent's goal is to reach $g_1$ with minimum total cost $C(\tau) = \sum_{t = 1 }^H c_t$, where $\tau = (s_1, d^s_1, a_1, \dots, s_H, d^s_H)$ is an execution of the task.\looseness=-1

We denote by $b_t^s$ a belief state---a representation of the history $(d^s_1, a_1, \ldots, d^s_t)$---and by $\gB$ the set of all belief states.
The agent maintains an \textit{\exepoltxt} $\exepol: \gB \times \gD \rightarrow \Delta(\gA)$.
The learning objective is to estimate an \exepoltxt that minimizes the expected total cost of performing tasks:
\begin{align}
    \min_{\pi} \mathbb{E}_{(s_1, g_1, d^g_1) \sim \mathfrak{T}, \tau \sim P_{\pi}(\cdot \mid s_1, d^g_1)}\left[ C(\tau) \right]
\label{eqn:standard_objective}
\end{align} where $P_{\pi}(\cdot \mid s_1, d^g_1)$ is the distribution over executions generated by a policy $\pi$ given start state $s_1$ and goal description $d^g_1$. 
In a standard POMDP, an agent performs tasks on its own, without asking for any external assistance.

%% file: sections/problem.tex
\section{Leveraging Assistance via Communication}

For an agent to accomplish tasks beyond its autonomous capabilities, we introduce an \textit{assistant}, who can provide rich information about the environment and the task.
We then describe how the agent requests and incorporates information from the assistant. 
\autoref{fig:illustration} illustrates an example communication between the agent and the assistant on an example object-finding task. 
Our formulation is inspired by the intention-based theory of human communication \citep{sperber1986relevance, tomasello2005understanding, scott2014speaking}, which characterizes communication as the expression and recognition of intentions in context.
In this section, we draw connections between the theory and elements of POMDP learning, which allows us to derive reinforcement learning algorithms to teach the agent to make intentional requests.\looseness=-1

\paragraph{Assistant.} 
We assume an ever-present assistant who knows the agent's current state $s_t$, the goal state $g_t$, and the optimal policy $\pi^{\star}$.
Their first capability is to provide a description of a state, defined by a function $\rho_A: \gS \times \gD \rightarrow \Delta(\gD)$ where $\rho_A(d' \mid s, d)$ specifies the probability of giving $d'$ to describe state $s$ given a current description $d$, which  can be empty.
The second capability is to propose a subgoal of a current goal, specified by a function $\omega_A: \gS \times \gS \rightarrow \Delta(\gS)$, where $\omega_A(g' \mid s, g)$ indicates the probability of proposing $g'$ as a subgoal given a current state $s$ and a goal state $g$.

\paragraph{Common ground.}
Common ground represents mutual knowledge between the interlocutors and is a prerequisite for communication \citep{clark1991grounding,stalnaker2002common}. 
In our context, knowledge of the agent is contained in its \exepoltxt $\exepol$, while knowledge of the assistant is given by $\pi^{\star}$.
When $\exepol$ and $\pi^{\star}$ maps to the same action distribution given an input $(b^s, d^g)$, we say that the input belongs to the common ground of the agent and the assistant. 
Substantial common ground is needed for effective communication to take place.
Thus, we assume that \exepoltxt has been pre-trained to a certain level of performance so that there exists a non-empty subset of inputs on which the outputs of $\exepol$ and $\pi^{\star}$ closely match.

\subsection{The \textit{Listener} Problem: \qquad\qquad\qquad\qquad Incorporating Rich Information Provided by the Assistant}

Upon receiving a request from the agent, the assistant replies with a state description $d \in \mathcal{D}$.
The generality of our notion of state description (see \autoref{sec:prelim}) means that the assistant can provide information in any medium and format, so long as it is compatible with the input interface of $\exepol$. 
This reflects that only information interpretable by the agent is useful. 

Concretely, the assistant can provide a new current-state description $d^s_{t + 1}$, which the agent appends to its history to compute a new belief state $b^s_{t + 1}$ (e.g., using a recurrent neural network).
The assistant can also provide a new goal description $d^g_{t + 1}$, in which case the agent simply replaces the current goal description $d^g_t$ with the new one. 
This formulation allows the human-agent communication protocol to be augmented in two ways: (i) enlarging the common ground by training $\exepol$ to interpret new state descriptions or (ii) designing the agent model architecture to support new types of input information. 
In comparison, frameworks that implement a reinforcement learning- or imitation learning-based communication protocol \citep[e.g.,][]{knox2009interactively, torrey2013teaching} allow humans to only give advice using low-bandwidth media like rewards or primitive actions.\looseness=-1

\subsection{The \textit{Speaker} Problem: \qquad\qquad\qquad\qquad Requesting Contextually Useful Information}
\label{sec:speaker_problem}

\paragraph{Asking Questions as a Cognitive Capability.} 
Asking questions is a cognitive process motivated by a person's self-recognition of knowledge deficits or  common ground mismatches \citep{graesser1992mechanisms}. 
The intention-based theory of human communication suggest that a question, like other communicative acts, should convey an information-seeking intention that is grounded in the speaking context.
In the case of question generation, the speaker's decision-making policy should be included in the context because, ultimately, a speaker asks a question to make better decisions.\looseness=-1

Approaches that teach an agent to mirror questions asked by humans \citep[e.g.,][]{de2017guesswhat},  
do not consider the agent's policy as part of the context for generating the question.
The questions selected by humans may not be useful for the learning agent. 
To address this issue, we endow the agent with the cognitive capability of anticipating how various types of information would affect its future performance.
The agent learns this capability using reinforcement learning, via \textit{interacting} with the assistant and the environment, rather than imitating human behaviors.

\paragraph{Information-Seeking Intentions.} 
While humans possess capabilities that help them come up with questions, it is unclear how to model those capabilities.
Some approaches \citep[e.g.,][]{mostafazadeh-etal-2016-generating,rao-daume-iii-2018-learning} rely on pre-composed questions, which are domain-specific.
We instead endow the agent with a set of intentions that correspond to speech acts in embodied dialogue \cite{thomason2020vision}.
They are relevant to solving general POMDPs, while being agnostic to the problem domain and the implementation of the execution policy:
\begin{enumerate}[itemsep=4pt]
    \item $\cur$: requests a new description of the current state $s_t$ and receives $d^s_{t + 1} \sim \rho_A\left(\cdot \mid s_t, d_t^s \right)$;
    \item $\goal$: requests a new description of the current goal $g_t$ and receives $d^{g}_{t + 1} \sim \rho_A\left(\cdot \mid g_{t}, d^g_t \right)$;
    \item $\sub$: requests a description of a subgoal and receives $d^{g}_{t + 1} \sim \rho_A\left(\cdot \mid g_{t + 1}, \emptyset\right)$ where the subgoal $g_{t + 1} \sim \omega_A\left(\cdot \mid s_{t}, g_{t}\right)$ and $\emptyset$ is an empty description.
\end{enumerate} 
We leave constructing more specific intentions (e.g., asking about a specific input feature) to future work.

\paragraph{Intention Selection.} To decide which intention to invoke, the agent learns an intention policy $\intpol$, which is itself a function of the execution policy $\exepol$. 
The intention policy's action space consists of five intentions: $\b \gA = \{\cur, \goal, \sub, \doa, \done\}$.
The first three actions 
convey the three intentions defined previously. 
The remaining two actions are used for traversing in the environment:
\begin{enumerate}[itemsep=4pt]
    \setcounter{enumi}{3}
    \item \doa: executes the most-probable action $a^{\textrm{do}}_{t} \triangleq \argmax_{a \in \gA} \hat \pi\left(a \mid b_{t}^s, d^g_{t}\right)$. The agent transitions to a new state $s_{t + 1} \sim T(s_t, a^{\textrm{do}}_{t})$;
    \item \done: decides that the current goal $g_{t}$ has been reached. 
    If $g_t$ is the main goal ($g_t = g_1$), the episode ends. If $g_t$ is a subgoal ($g_t \neq g_1$), the agent may choose a new goal to follow. 
\end{enumerate}

In our implementation, we feed the hidden features and the output distribution of $\exepol$ as inputs to $\intpol$ so the agent can use its execution policy in choosing its intentions. 
Later we will specify the input space of the interaction policy and how the next subgoal is chosen (upon \done), as these details depend on how the agent implements its goal memory. 
The next section introduces an instantiation where the agent uses a stack data structure to manage the goals it receives. 

%% file: sections/framework.tex
\section{Learning When and What to Ask}

In this section, we formalize the \framework framework for learning to select information-seeking intentions.
We construct the POMDP environment that the intention policy acts in, referred to as the \textit{\intenvtxt} (\autoref{sec:interaction_environment}) to distinguish with the environment that the \exepoltxt acts in. 
Our construction employs a \textit{goal stack} to manage multiple levels of (sub)goals  (\autoref{sec:goal_stack}).
A goal stack stores all the (sub)goals the agent has been assigned but has not yet decided to terminate.
It is updated in every step depending on the selected intention.
In \autoref{sec:cost_function}, we design a cost function that trades off between taking few actions and completing tasks.

\subsection{Intention Environment}
\label{sec:interaction_environment}

Given an \exeenvtxt $E = (\gS, \gA, T, c, \gD, \rho)$, the \intenvtxt is a POMDP $\b E = (\b \gS, \b \gA, \b \gT, \b c, \b \gD, \b \rho)$: 
\begin{itemize}[leftmargin=*,itemsep=4pt]
    \item \textbf{State space} $\b \gS = \gS \times \gD \times \gG_L$, where $\gG_L$ is the set of all goal stacks of at most $L$ elements. Each state $\b s = (s, d^s, G) \in \b \gS$ is a tuple of an execution state $s$, description $d^s$, and goal stack $G$. Each element in the goal stack $G$ is a tuple
    of a goal state $g$ and description $d^g$;\looseness=-1
    \item \textbf{Action space} $\b \gA = \{ \cur, \goal, \sub, \doa, \done \}$;
    \item \textbf{State-transition} function $\b T = T_s \cdot T_G$ where $T_s : \gS \times \gD \times \b \gA \rightarrow \Delta(\gS \times \gD)$ and $T_G: \gG_L \times \b \gA \rightarrow \Delta(\gG_L)$;
    \item \textbf{Cost function} $\b c: (\gS \times \gG_L) \times \b \gA \rightarrow \mathbb{R}$, defined in \autoref{sec:cost_function} to trade off operation cost and task error;\looseness=-1
    \item \textbf{Description space} $\b \gD = \gD \times \gG^d_L$ where $\gG^d_L$ is the set of all goal-description stacks of size $L$.
    The agent cannot access the environment's goal stack $G$, which contains true goal states, but only observe descriptions in $G$. We call this partial stack a \textit{goal-description stack}, denoted by $G^d$;
    \item \textbf{Description function} $\b \rho : \b \gS \rightarrow \b \gD$, where $\b \rho(\b s) = \b\rho(s, d^s, G) = (d^s, G^d)$. Unlike in the standard POMDP formulation, this description function is deterministic.
\end{itemize}

A belief states $\b b_t$ of the \intenvtxt summarizes a history $(\b s_1, \b a_1, \cdots, \b s_t)$. 
We define the \intpoltxt as $\inpol: \b \gB \rightarrow \Delta(\gA)$, where $\b \gB$ is the set of all belief states. 

\subsection{Goal Stack}
\label{sec:goal_stack}

The goal stack is a list of tasks that the agent has not declared complete. 
The initial stack $G_1 = \left\{ \left(g_1, d^g_1\right) \right\}$ contains the main goal and its description.
At time $t$, the agent
executes the goal at the top of the current stack, i.e. $g_t = G_t.\textrm{top}()$.
Only the \goal, \sub, and \done actions alter the stack. 
\goal replaces the top goal description of $G_t$ with the new description $d^g_{t + 1}$ from the assistant. 
\sub, which is only available when the stack is not full, pushes a new subtask $(g_{t + 1}, d^g_{t + 1})$ to $G_t$.
\done pops the top element from the stack.
The goal-stack transition function is $T_G(G_{t + 1} \mid G_t, \b a_t) = \mathds{1}\{G_{t + 1} = G_t.\texttt{update}(\b a_t)\}$ where $\mathds{1}\{.\}$ is an indicator and $G_t.\textrm{update}(a)$ is the updated stack after action $a$ is taken. 

\subsection{State Transition}

The transition function $T_s$ is factored into two terms:
\begin{align}
    &T_s(s_{t + 1}, d^s_{t + 1} \mid  s_t, d^s_t, \b a_t) = \nonumber \\ &\qquad\qquad\underbrace{P(s_{t + 1} \mid s_t, \b a_t)}_{\triangleq p_{\textrm{state}}(s_{t + 1})} \cdot \underbrace{P(d^s_{t + 1} \mid s_{t + 1}, d^s_t, \b a_t)}_{\triangleq p_{\textrm{desp}}(d^s_{t + 1})}
\end{align} 
Only the \doa action changes the execution state, with: $p_{\textrm{state}}(s_{t + 1})  = T\left( s_{t + 1} \mid s_t, a^{\textrm{do}}_t \right)$, where $a^{\textrm{do}}_t$ is the action chosen by $\exepol$ and $T$ is the \exeenvtxt's state transition function.
The current-state description is altered when the agent requests a new current-state description (\cur), where $p_{\textrm{desp}}(d^s_{t + 1}) = \rho_A(d^s_{t + 1} \mid s_{t + 1}, d^s_t)$, or moves (\doa), where $p_{\textrm{desp}}(d^s_{t + 1}) = \rho(d^s_{t + 1} \mid s_{t + 1})$ (here, $\rho$ is the description function of the \exeenvtxt).

\subsection{Cost Function}
\label{sec:cost_function}

The \intpoltxt needs to trade-off between two types of cost: the cost of operation (making information requests and traversing in the environment), and the cost of not completing a task (task error). 
The two types of cost are in conflict: the agent may lower its task error if it is willing to suffer a larger operation cost by making more requests to the assistant.

We employ a simplified model where all types of cost are non-negative real numbers of the same unit. 
Making a request of type $a$ is assigned a constant cost $\gamma_a$.
The cost of taking the \doa action is $c(s_t, a^{\textrm{do}}_t)$, the cost of executing the $a^{\textrm{do}}_t$ action in the 
environment.
Calling \done to terminate execution of the main goal $g_1$ incurs a task error $c(s_t, a_{\textrm{done}})$.
We exclude the task errors of executing subgoals because the \intpoltxt is only evaluated on reaching the main goal.
The cost function is concretely defined as follows
\begin{align}
    \b c(s_t, G_t, \b a_t) = \begin{cases}
    c(s_t, g_t, a^{\textrm{do}}_t) &\text{if $\b a_t = \doa$}, \\
    \gamma_{\b a_t} &\text{if $\b a_t \notin \left\{ \doa, \done \right\}$}, \\
    c(s_t, g_t, a_{\textrm{done}}) &\text{if $\b a_t = \done, |G_t| = 1$}, \\
    0 &\text{if $\b a_t = \done, |G_t| > 1$} \\
    \end{cases} \nonumber
\end{align} 
The magnitudes of the costs naturally specify a trade-off between operation cost and task error. 
For example, setting the task errors much larger than the other costs indicates that completing tasks is prioritized over taking few actions. 

To facilitate learning, we apply reward shaping \citep{ng1999policy}, augmenting a cost $c(s, G, a)$ with $\Phi(s, g)$, the task error received if the agent terminated in $s$.
The cost received at time step $t$ is $\tilde c_t \triangleq \b c_t + \Phi(s_{t + 1}, g_{t + 1}) - \Phi(s_{t}, g_{t})$.
We assume that after the agent terminates the main goal, it transitions to a special terminal state $s_{\textrm{term}} \in \gS$ and remains there.
We set $\Phi(s_{\textrm{term}}, \texttt{None}) = 0$, where $g_t = \texttt{None}$ signals that the episode has ended. 
The cumulative cost of an execution under the new cost function is \looseness=-1
\begin{align}
    \sum_{t = 1}^{H} \tilde c_t 
    &= \sum_{t = 1}^{H} \Big[ \b c_t + \Phi(s_{t + 1}, g_{t + 1}) -  \Phi(s_{t}, g_{t}) \Big] \\
    &= \sum_{t = 1}^{H} \b c_t - \Phi(s_{1}, g_{1})
\end{align} Since $\Phi(s_{1}, g_{1})$ does not depend on the action taken in $s_1$, minimizing the new cumulative cost does not change the optimal policy for the task $(s_1, g_1)$. 

%% file: sections/exp_setup.tex
\section{
Modeling Human-Assisted Navigation}
\label{sec:han}

\paragraph{Problem.}
We apply \framework to modeling a human-assisted navigation (\han) problem, in which a human requests an agent to find an object in an indoor environment. 
Each task request asks the agent to go to a room of type $r$ and find an object of type $o$ (e.g., find a mug in a kitchen). 
The agent shares its current view with the human (e.g. via an app).
We assume that the human is 
familiar with the environment and 
can recognize the agent's location.
Before issuing a task request, the human imagines a goal location (not revealed to the agent).
We are 
interested in evaluating success in  \textit{goal-finding}, i.e. whether the agent can arrive at the human's intended goal location. 
Even though there could be multiple locations that match a request, the agent only succeeds if it arrives exactly at the chosen goal location.

\paragraph{Environment.}
We construct the execution environments using the house layout graphs provided by the Matterport3D simulator \citep{anderson2018vision}.
Each graph is generated from a 3D model of a house where each node is a location in the house and each edge connects two nearby unobstructed locations. 
At any time, the agent is at a node of a graph.
Its action space $\gA$ consists of traversing to any of the nodes that are adjacent to its current node.

\paragraph{Scenario.}
We simulate the scenario where the agent is pre-trained in simulated environments and then deployed in real-world environments. 
Due to mismatches between the pre-training and real-world environments, the capability of the agent degrades at deployment time.
It may not reliably recognize
objects, the room it is currently in, or the intent of a request. 
However, a simulated human assistant is available to provide additional information about the environment and the task, which helps the agent relate its current task to the ones it has learned to fulfill during pre-training.

\paragraph{Subgoals.}
If the agent requests a subgoal, the assistant describes a new goal roughly halfway to its current goal (but limited to a maximum distance $l_{\textrm{max}}$).
Specifically, let $p$ be the shortest path from the agent's current state $s_t$ to the current goal $g_t$, and $p_i$ be the $i$-th node on the path ($0 \leq i < |p|$). 
The subgoal location is chosen as $p_{k}$ where $k = \min(\floor{|p| / 2}, l_{\textrm{max}})$, where $l_{\textrm{max}}$ is a pre-defined constant.\looseness=-1

\paragraph{State Description.} We employ a discrete bag-of-features representation for state descriptions.\footnote{While our representation of state descriptions simplifies the object/room detection problem for the agent, it does not necessarily make the navigation problem easier than with image input, as images may contain information that is not captured by our representation (e.g., object shapes and colors, visualization of paths).} 
A bag-of-feature description ($d^s$ or $d^g$) emulates the information that the agent has extracted from a raw input that it perceives (e.g., an image, a language sentence).
Concretely, we assume that when the agent sees a view, it detects the room and nearby objects.
Similarly, when it receives a task request or a human response to its request, it identifies information about rooms, objects, and actions in the response. 
Working with this discrete input representation allows us to easily simulate various types and amounts of information given to the agent.

Specifically, we model three types of input features: the name of a room, information about an object (name, distance and direction relative to a location), and the description of a navigation action (travel distance and direction). 
The human can supply these types of information to assist the agent.
We simulate two settings of descriptions: \textit{dense} and \textit{sparse}.
A dense description is a variable-length lists containing the following features: the current room's name and features of at most 20 objects within five meters of a viewpoint. 
Sparse descriptions represent imperfect perception of the agent in real-world environments. 
A sparse description is derived by first constructing a dense description and then removing the features of objects that are not in the top 100 most frequent (out of $\sim$1K objects).
The sparse description of the current location ($d^s$) does not contain the room name, but that of the goal location ($d^g$) does.
As each description is a sequence of feature vectors, whose length varies depend on the location,
we use a Transformer model \citep{vaswani2017attention} to be able to encode such inputs into continuous feature vectors. 
Details about the feature representation and the model architecture are provided in the Appendix.

\looseness=-1

\paragraph{Experimental Procedure.} 
We conduct our experiments in three phases.
In the \textit{pre-training} phase, the agent learns an \exepoltxt $\oppol$ with access to only dense descriptions. 
This emulates training in a simulator that supplies rich information to the agent. 

In the \textit{training} phase, the agent is only given sparse descriptions of its current and goal locations. 
This mimics the degradation of the agent's perception about the environment and the task when deployed in real-world conditions.
In this phase, the agent can request dense descriptions from the human. 
Upon a request for information about a (goal or current) location, the human gives a dense description with room and object features of that location.
However, when the agent requests a subgoal (\sub) \textit{and} is adjacent to the subgoal that the human wants to direct it to, the human simply gives a description of the next optimal navigation action to get to that location.
We use advantage actor-critic \citep{mnih2016asynchronous} to learn an \intpoltxt $\inpol$ that determines which type of information to request. 
The \intpoltxt is now trained in environment graphs that are previously seen as well as unseen during the pre-training phase.

Finally, in the \textit{evaluation} phase, the agent is tested on three conditions: seen environment and target object type but starting from a new room (\unseenstr), seen environment but new target object type (\unseenobj), and unseen environment (\unseenenv). 
The execution policy $\oppol$ is fixed during the training and evaluation phases.
We created 82,104 examples for pre-training, 65,133 for training, and approximately 2,000 for each validation or test set.
Additional details about the training procedure and dataset are included in the Appendix.

%% file: sections/results.tex
\begin{table*}[t!]
\centering
\footnotesize
\caption{Test goal success rates 
and 
the average number of different types of actions taken by the agent (on all task types).} 

\begin{tabular}{@{}lcccc@{\hspace{2pt}}c@{\hspace{2pt}}c@{\hspace{2pt}}c@{}}

    \toprule
    & \multicolumn{3}{c}{Success Rate \% $\uparrow$} & \multicolumn{4}{c@{}}{Avg. number of actions $\downarrow$}   \\
    \multicolumn{1}{@{}c}{} & Unseen & Unseen & Unseen &  &  &  &  \\
    \multicolumn{1}{@{}c}{Agent} & Start & Object & Environment & \cur & \goal & \sub & \doa \\
    \cmidrule(r){1-1} \cmidrule(lr){2-4} \cmidrule(lr){5-8}
    \multicolumn{7}{@{}l}{\textbf{Rule-based \intpoltxt $\inpol$} \ti{when to call \done is decided by the \exepoltxt $\exepol$}} \\
    \multicolumn{7}{@{}l}{ \ti{$d^s$: current-state description, $d^g$: goal description}} \\
    \quad No assistance \ti{always \doa until \done} & 43.4 & 16.4 & ~~3.0 & - & - & - & 13.1 \\
    \quad Dense $d^g$ \ti{\goal then always \doa until \done} & 67.2 & 56.6 & ~~9.7 & - & 1.0 & - & 12.6 \\
    \quad Dense $d^s$ \ti{always \cur then \doa until \done} & 77.9 & 30.6 & ~~4.1 & 12.0 & - & - & 12.0 \\
    \quad Dense $d^g$ and $d^s$ \ti{\goal then always \cur then \doa until \done}& 97.8 & 81.7 & ~~9.4 & 11.0 & 1.0 & - & 11.0 \\
    \quad Random + rules to match with \# of actions of learned-RL $\intpol$  & 78.8 & 68.5 & 12.7 & ~~2.0 & 1.0 & 1.7 & 11.3 \\
    \midrule
    \multicolumn{7}{@{}l}{\textbf{learned-RL \intpoltxt $\inpol$}} \\
    \quad With pre-trained navigation policy $\hat{\pi}$ (ours) & 85.8 & 78.2 & 19.8 & 2.1 & 1.0 & 1.7 & 11.1 \\
    \quad With uncooperative assistant \ti{change a request randomly to \cur, \goal or \sub} & 81.1 & 71.2 & 16.1 & 2.7 & 2.7 & 1.5 & ~~9.6 \\
    \quad With perfect navigation policy on sub-goals (skyline) & 94.3 & 95.1 & 92.6 & 0.0 & 0.0 & 6.3 & ~~7.3 \\
    \bottomrule
\end{tabular}
\label{tab:main_results}
\end{table*}

\section{Results and Analyses}

\paragraph{Settings.} In our main experiments, we set: the cost of taking a \cur, \goal, \sub, or \doa action to be 0.01 (we will consider other settings subsequently), the cost of calling \done to terminate the main goal (i.e. task error) equal the (unweighted) length of the shortest-path from the agent's location to the goal, and the goal stack's size ($L$) to be 2. 

We compare our learned-RL \intpoltxt with several rule-based \textit{baselines}.
Their descriptions are given in \autoref{tab:main_results}.
We additionally construct a strong baseline that first takes the \goal action (to clarify the human's intent) and then selects actions in such a way to match the distribution of actions taken by our RL-trained policy.
In particular, this policy is constrained to take at most $\floor{X_a} + y$ actions of type $a$, where $y \sim \textrm{Bernoulli}(X_a - \floor{X_a})$ and $X_a$ is a constant tuned on the validation set so that the policy has the same average count of each action as the learned-RL policy.
To prevent early termination, we enforce that the rule-based policy cannot take more \done actions than \sub actions unless  its \sub action's budget is exhausted.
When there is no constraint on taking an action, the policy chooses a random action in the set of available actions. 
Our learned-RL policy can only outperform this policy by being able to determine contextually useful information to request, which is exactly the capability we wish to evaluate. 
We also construct a \textit{skyline} where the \intpoltxt is also learned by RL but with an \exepoltxt that always fulfill subgoals perfectly. \looseness=-1

\begin{figure*}[t!]
    \begin{minipage}{0.5\linewidth}
    \centering
    \begin{subfigure}[t]{0.49\linewidth}
        \centering
        \includegraphics[width=\linewidth,clip,trim=6 6 14 2]{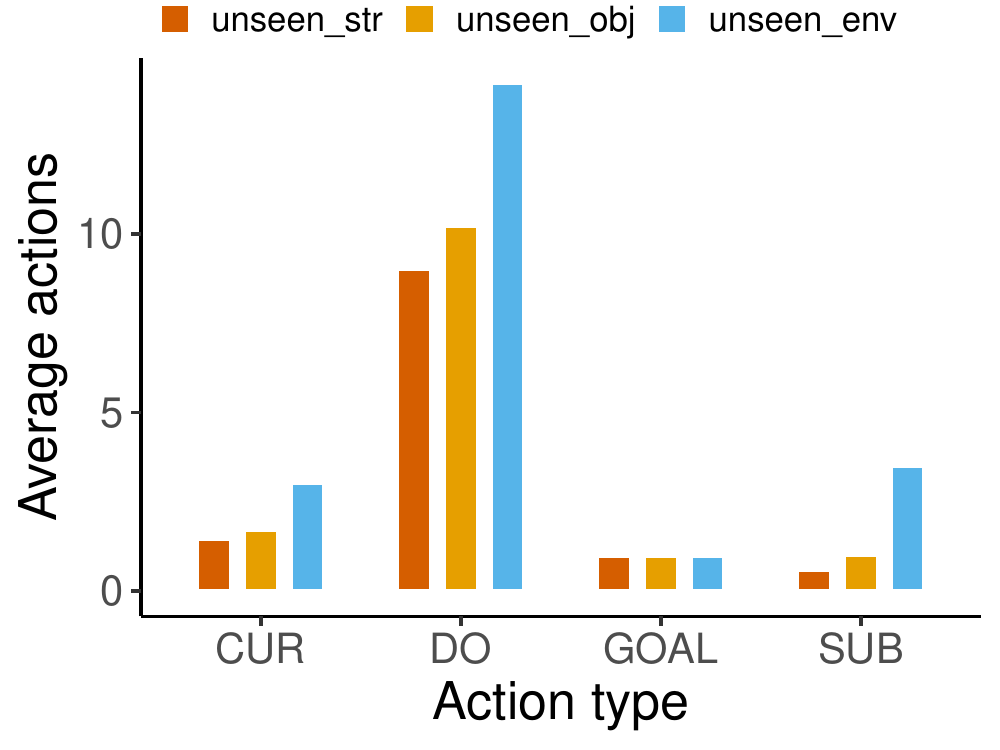}
        \captionsetup{width=.95\linewidth,font=footnotesize}
        \caption{
        Subgoals are requested much more in unseen environments.}
        \label{fig:action_vs_env}
    \end{subfigure}
    \begin{subfigure}[t]{0.5\linewidth}
        \centering
        \includegraphics[width=\linewidth,clip,trim=5 6 32 2]{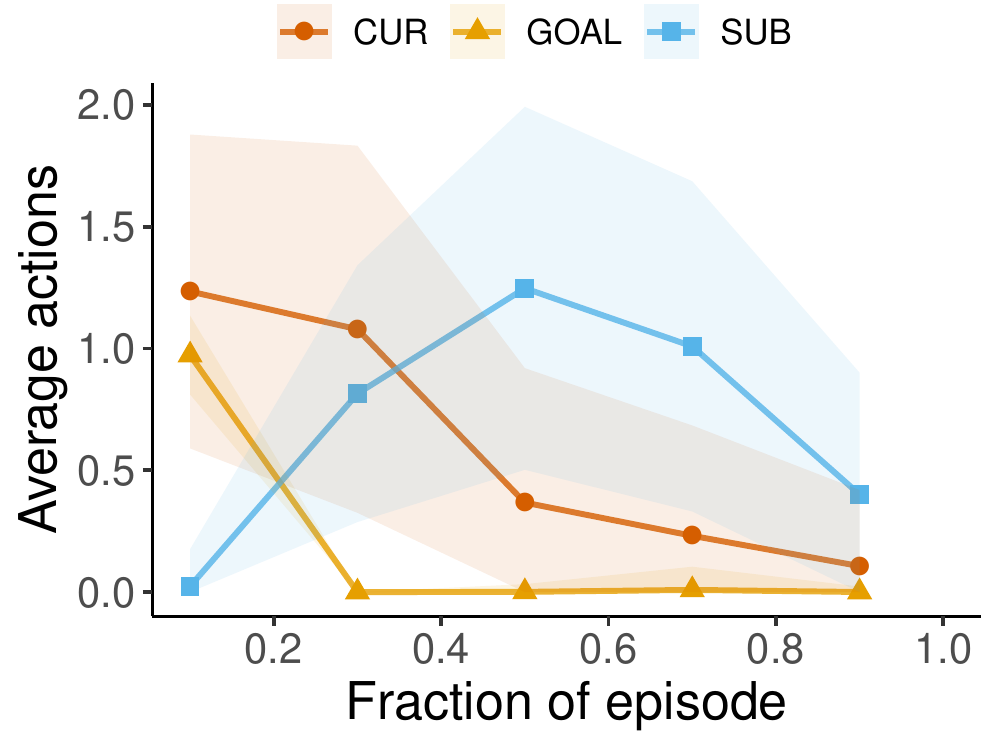}
        \captionsetup{width=.95\linewidth,font=footnotesize}
        \caption{
        Subgoals are requested in the middle, goal information at the beginning.
        }
        \label{fig:action_over_time}
    \end{subfigure}
    \captionsetup{width=.9\linewidth}
    \caption{Analyzing the behavior of the learned-RL \intpoltxt (on validation environments).}
    \end{minipage}
    \centering
    \begin{minipage}{0.49\linewidth}
    \begin{subfigure}[t]{0.49\linewidth}
        \centering
        \includegraphics[width=\linewidth]{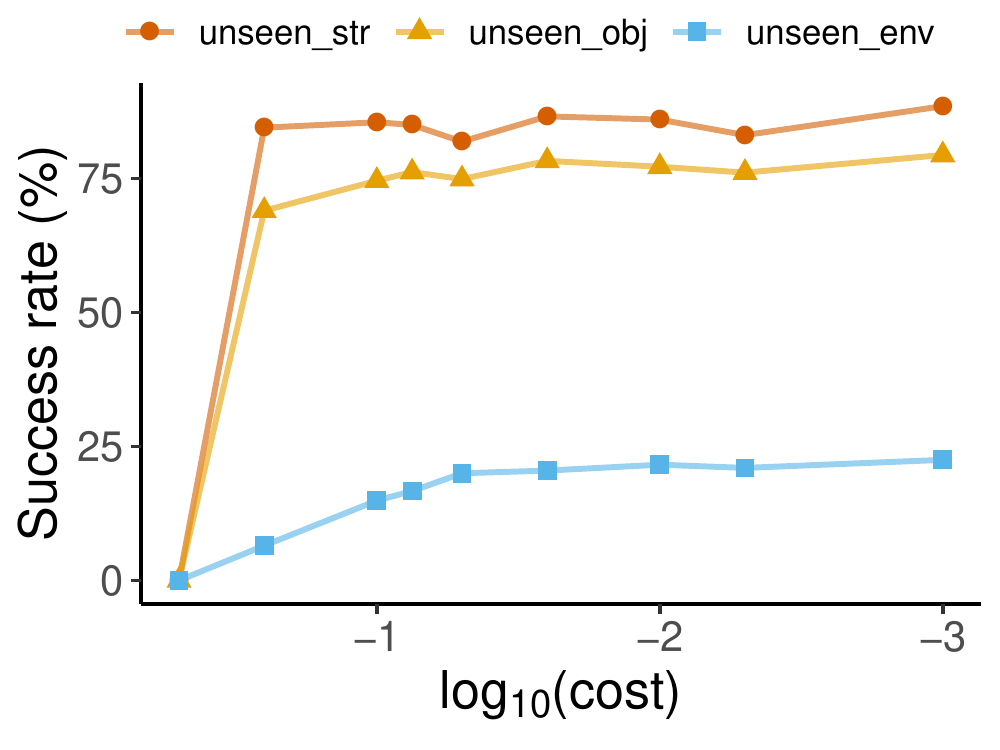}
        \captionsetup{width=.95\linewidth,font=footnotesize}
        \caption{Effect of cost on success.}
        \label{fig:action_cost_vs_perf}
    \end{subfigure}
    \begin{subfigure}[t]{0.5\linewidth}
        \centering
        \includegraphics[width=\linewidth]{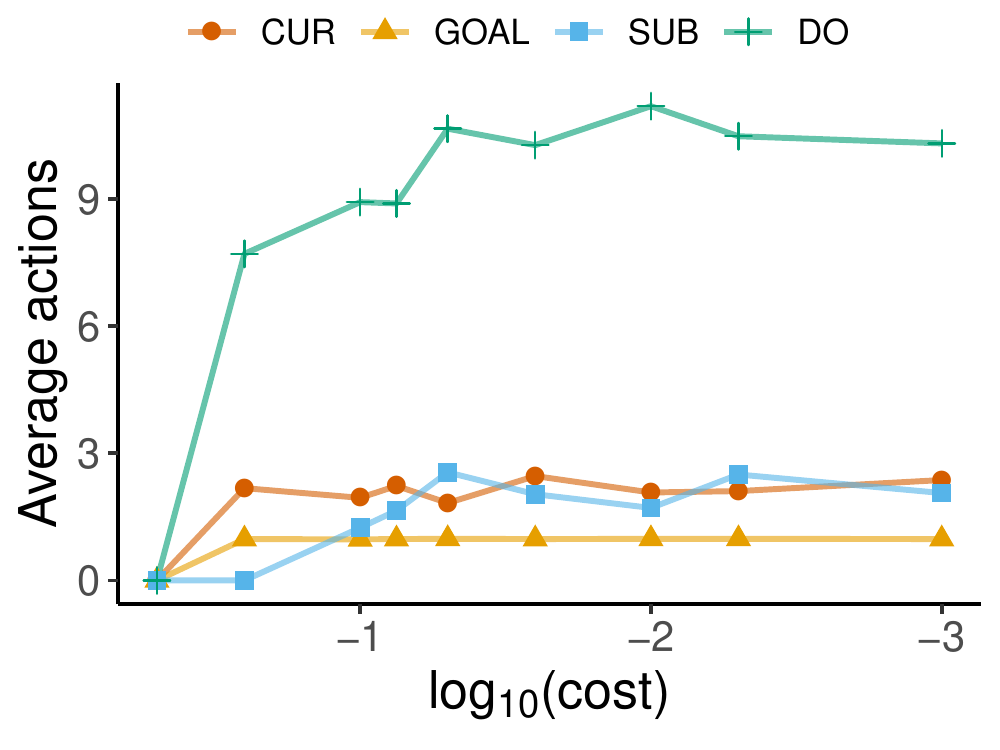}
        \captionsetup{width=.95\linewidth,font=footnotesize}
        \caption{Effect of cost on actions.}
        \label{fig:action_cost_vs_action_dist}
    \end{subfigure}
    \captionsetup{width=.9\linewidth}
    \caption{Analyzing the effect of simultaneously varying the cost of the \cur, \goal, \sub, \doa actions (on validation environments), thus trading off success rate versus number of actions taken.}
    \label{fig:hyper}
    \end{minipage}
\end{figure*}

\paragraph{In the problem we construct, different types of information are useful in different contexts (\autoref{tab:main_results}).} 
Comparing the rule-based baselines reveals the benefits of making each type of request. 
Overall, we observe that information about the current state is helpful on only tasks in seen environments (\unseenstr and \unseenobj). 
Information about the goal greatly improves performance of the agent in unseen environments (dense-$d^g$ outperforms no-assistance by 6.4\% in \unseenenv).
Dense-$d^g$ outperforms dense-$d^s$ in \unseenobj but underperforms in \unseenstr, showing that goal information is more useful than current-state information in finding new objects but less useful in finding seen ones. 
The two types of information is complementary: dense-$d^g$-and-$d^s$ improves success rate versus dense-$d^g$ and dense-$d^s$ in most evaluation conditions. 

We expect subgoal information to be valuable mostly in unseen environments. 
This is confirmed by the superior performance of our learned-RL policy, which can request subgoals, over the rule-based baselines, which do not have this capability, in those environments. 

\paragraph{Our learned-RL policy learns the advantages of each type of information (\autoref{tab:main_results} \& \autoref{fig:action_vs_env}).} 
Aided by our learned-RL policy, the agent observes a substantial $\sim$2$\times$ increase in success rate on \unseenstr, $\sim$5$\times$ on \unseenobj, and $\sim$7$\times$ on \unseenenv, compared to when performing tasks without assistance. 
This result indicates that the assistance-requesting skills are increasingly more helpful as the tasks become more unfamiliar.
\autoref{fig:action_vs_env} shows the request pattern of the policy in each evaluation condition. 
Our policy learns that it is not useful to make more than one \goal request. 
It relies increasingly on \cur and \sub requests in more difficult conditions (\unseenobj and \unseenenv). 
The policy makes about 3.5$\times$ more \sub requests in 
\unseenenv than in \unseenobj.
Specifically, with the capability of requesting subgoals, the agent impressively doubles the success rate of the dense-$d^g$-and-$d^s$ policy in unseen environments, which \textit{always} has access to dense descriptions in these environments. 
Moreover, our policy does not have to bother the assistant in every time step: only $\nicefrac{1}{4}$ of its actions are requests for information.

\paragraph{Our learned-RL policy selects contextually useful requests (\autoref{tab:main_results}, bottom half \& \autoref{fig:action_over_time}).  } 
The policy is significantly more effective than the rule-based baseline that uses the same number of average actions per episode (+7.1\% on \unseenenv).  
Note that we enforce rules so that this baseline only differs from our learned-RL policy in where it places \cur and \sub requests. 
Thus, our policy has gained advantages by making these requests in more appropriate situations. 
To further showcase the importance of being able to obtain the right type of information, we use RL to train a policy with an \textit{uncooperative} assistant, who disregards the agent's request intention and instead replies to an intention randomly selected from $\{\cur, \goal, \sub\}$.
As expected, performance of the agent drops in all evaluation conditions. 
This shows that our agent has effectively leveraged the cooperative assistant by making useful requests. 

We also observe that the agent adapts its request strategy through the course of an episode.
As observed in \autoref{fig:action_over_time}.
the \goal action, if taken, is always taken only once and immediately in the first step.
The number of \cur actions gradually decreases over time. 
The agent makes most \sub requests in the middle of an episode, after its has attempted but failed to accomplish the main goals. 
We observe similar patterns on the other two validation sets.

Finally, results obtained by the skyline shows that further improving performance of the \exepoltxt on short-distance goals would effectively enhance the agent's performance on long-distance goals.

\paragraph{Effects of Varying Action Cost (\autoref{fig:hyper}).}  
We assign the same cost to each \cur, \goal, \sub, or \doa action. 
\autoref{fig:action_cost_vs_perf} demonstrates the effects of changing this cost on the success rate of the agent. 
An equal cost of 0.5 makes it too costly to take any action, inducing a policy that always calls \done in the first step and thus fails on all tasks.
Overall, the success rate of the agent rises as we reduce the action cost.
The increase in success rate is most visible in \unseenenv and least visible in \unseenstr.
Similar patterns are observed with various time limits ($H = 10, 20, 30$).
\autoref{fig:action_cost_vs_action_dist} provides more insights. 
As the action cost decreases, we observe a growth in the number of \sub and \doa actions taken by the \intpoltxt. 
Meanwhile, the numbers of \cur and \goal actions are mostly static. 
Since requesting subgoals is more helpful in unseen environments, the increase in the number of \sub actions leads the more visible boost in success rate on \unseenenv tasks.

\begin{table}[t!]
\setlength{\tabcolsep}{2pt}
\centering
\footnotesize
\caption{Success rates and numbers of actions taken with different stack sizes (on validation). Larger stack sizes significantly aid success rates in unseen environments, but not in seen environments.}

\begin{tabular}{c@{~}lccccccc}
    \toprule
    && \multicolumn{3}{c}{Success rate (\%) $\uparrow$} & \multicolumn{4}{c}{Average \# of actions $\downarrow$} \\
    \multicolumn{2}{c}{Stack} & Unseen & Unseen & Unseen &  &  &  & \\
    \multicolumn{2}{c}{size} & Start & Obj. & Env. & \cur & \goal & \sub & \doa\\
    \midrule
    1 & \ti{no subgoals} & 92.2 & 78.4 & 12.5 & 5.1 & 1.9 & 0.0 & 10.7 \\
    2 & & 86.9 & 77.6 & 21.6 & 2.1 & 1.0 & 1.7 & 11.2 \\
    3 & & 83.2 & 78.6 & 33.5 & 1.3 & 1.0 & 5.0 & 8.2 \\
    \bottomrule
\end{tabular}
\label{tab:deeper_stack}
\end{table}

\paragraph{Recursively Requesting Subgoals of Subgoals (\autoref{tab:deeper_stack}).} 
In \autoref{tab:deeper_stack}, we test the functionality of our framework with a stack size 3, allowing the agent to request subgoals of subgoals. 
As expected, success rate on \unseenenv is boosted significantly (+11.9\% compared to using a stack of size 2). 
Success rate on \unseenobj is largely unchanged; we find that the agent makes more \sub requests on those tasks (averagely 4.5 requests per episode compared to 1.0 request made when the stack size is 2), but doing so does not further enhance performance. 
The agent makes less \cur requests, possibly in order to offset the cost of making more \sub requests, as we keep the action cost the same in these experiments. 
Due to this behavior, success rate on \unseenstr declines with larger stack sizes, as information about the current state is more valuable for these tasks than subgoals. 
These results show that the critic model overestimates the $V$ values in states where $\sub$ actions are taken, leading to the agent learning to request more subgoals than needed. 
This suggests that the our critic model is not expressive enough to encode different stack configurations. 

%% file: sections/related.tex
\section{Related Work}

\paragraph{Knowledge Transfer in Reinforcement Learning.} 
Various frameworks have been proposed to model information transfer between humans and agents \citep{da2019survey}. 
A basic human-agent communication protocol is instruction-following \citep{chen2011learning,bisk2016natural,alomari2017natural,anderson2018vision,misra-etal-2018-mapping,yu2018guided}, where agents execute a natural language request. 
For multiple-round communication, \citet{torrey2013teaching} introduce the action-advising framework where a learner decides situations where it needs to request reference actions from a teacher.
This can be viewed as active learning \citep{angluin1988queries,cohn1994improving} in an RL setting.
The request policy can be constructed based on uncertainty measurements 
\citep{da2020uncertainty,culotta2005reducing} or learned via RL \citep{fang2017learning}.
An agent-to-agent variant poses the problem of learning a teaching policy in addition to the request policy \citep{da2017simultaneously,zimmer2014teacher,omidshafiei2019learning}. 
Overall, this literature uses IL-based communication protocols that assume the teacher shares a common action space with the learner and provides actions in this space in response to the agent's requests.
Others employ standard RL communication protocols, where the human conveys knowledge through numerical scores or categorical feedback \citep{knox2009interactively,judah2010reinforcement,griffith2013policy,peng2016need,christiano2017deep,lee2021pebble}.
In \citet{maclin1996creating}, 
humans advise the agent with domain-specific language, specifying rules to incorporate into the agent's model.

Recent frameworks \citep{jiang2019language,kim2019learning,nguyen2019vnla,nguyen2019hanna} allow the teacher to specify high-level goals instead of low-level actions or rewards. 
\framework generalizes these, allowing specification
of subgoal information and descriptions of the current and goal states. 
Importantly, the framework enables the agent to convey specific 
intentions rather than passively receiving instructions or calling for generic help.

Information expressed in human language has been incorporated into RL frameworks to guide task execution \citep{hermann2017grounded} or assist with learning \citep{ammanabrolu2020avoid}---see \citet{luketina2019survey} for a thorough survey.
Language-based inverse RL frameworks infer the reward function from language utterances
\citep{fu2019language,sumers2020learning, akakzia2021grounding,zhou2021inverse}.
In contrast, we investigate human-agent communication \textit{during} task execution---the agent does \textit{not} update its task-executing policy. 
Similar to language-conditioned RL, we directly incorporate feedback as input to the agent's policy. 
This approach also bears a resemblance to work on learning from language description \citep{jiang2019language,chan2019actrce,colas2020language,cideron2020higher,nguyen2021iliad}, except that the agent does not learn from the response and can incorporate more diverse feedback. 
Our setting is closely related to \citet{he2013dynamic}, where an agent 
selects a subset of useful input features, but we remove the assumption that features arrive in a fixed order. 

Our work requires solving a hierarchical reinforcement learning problem \citep{sutton1999between,kulkarni2016hierarchical,le2018hierarchical,chane2021goal}. 
While standard formulations mostly consider two levels of goal, our stack-based implementation offers a convenient way to construct deeper goal hierarchies.

\paragraph{Task-Oriented Dialog and Natural Language Questions.} 
\framework models a task-oriented dialog problem. 
Most problem settings require the agent to compose questions to gather information from humans \citep{de2017guesswhat,das2017visual,de2018talk,thomason2020vision,padmakumar2022teach,narayan-chen-etal-2019-collaborative,shi2022learning}. 
A related line of work is generating language explanations of model decisions \citep{camburu2018snli,hendricks2016generating,rajani2019explain}.
The dominant approach in these spaces is to mimic pre-collected human utterances.
However, naively mirroring human behavior cannot enable agents to understand the limits of their knowledge. 
We take a different approach, teaching the agent to understand its intrinsic needs through interaction.
Teaching agents to act and communicate with intentions is understudied \citep{karimpanal2021intentionality}.
Notable work in this space proposes a computational theory-of-mind for agents to enable them to reason about their own mental states and those of others \citep{andreas-klein-2016-reasoning,fried-etal-2018-unified,kang2020incorporating,roman-roman-etal-2020-rmm,zhu2021few,bara-etal-2021-mindcraft}.
Our work is not concerned about designing theory-of-mind models, instead concentrating on highlighting the importance of interaction in the learning of intentional behavior.

%% file: sections/conclusion.tex
\section{Conclusion}
Our framework can theoretically capture rich human-agent communication, and even communication with other knowledge sources like search engines or language models \citep{liu2022asking}. 
An important empirical question is how well it generalizes to more realistic interactions and environments (e.g., \citep{shridhar2020alfred}).
Especially when deploying with real humans, using human simulations to pre-train the agent can help reducing human effort. 
Large language models \citep{brown2020language,rae2021scaling,chowdhery2022palm} can be leveraged for constructing such simulators. 
Towards generating more specific, natural questions, methods for generating faithful explanations \citep{kumar-talukdar-2020-nile,madsen2021evaluating}, measuring feature importance \citep{zeiler2014visualizing,koh2017understanding,das2020opportunities}, or learning the casual structure of black-box policies \citep{geiger2021inducing} are relevant to investigate. 
Another exciting direction is to enable the agent to learn from the obtained information. 
We expect that, similar to curriculum-based learning \citep{wang2019paired}, the communication protocol of our framework would guide the agent to request and learn increasingly more difficult tasks over time.

%% file: sections/appendix_content.tex
\section{Training Procedure}
\label{sec:han_method}

\paragraph{Training Algorithms.} 
We pre-train the execution policy $\exepol$ with DAgger \citep{ross2011reduction}, minimizing the cross entropy between its action distribution with that of a shortest-path oracle (which is a one-hot distribution with all probability concentrated on the optimal action). 

We use advantage actor-critic \citep{mnih2016asynchronous} to train the \intpoltxt $\inpol$. 
This method simultaneously estimates an actor policy $\inpol : \b \gB \rightarrow \Delta(\b \gA)$ and a critic function $\vfunc : \b \gB \rightarrow \mathbb{R}$.
Given an execution $\b \tau = (\b s_1, \b a_1, \b c_1 \cdots, \b s_H)$, the gradients with respect to the actor and critic are
\begin{align}
    \nabla_{\theta} \mathcal{L}_{\textrm{actor}} 
    &= \sum_{t = 1}^H \left( \vfunc(\b b_t^v) - C_t \right)\nabla_{\theta} \log \inpol(\b a_t \mid \b b_t^a) \\
    \nabla_{\phi} \mathcal{L}_{\textrm{critic}} 
    &= \sum_{t = 1}^H \left(\vfunc(\b b_t^v) - C_t \right) \nabla_{\phi} \vfunc (\b b_t^v)
\end{align} where $C_t = \sum_{j = t}^H c_j$,  $\b b_t^a$ is a belief state that summarizes the partial execution $\b \tau_{1:t}$ for the actor, and $\b b_t^v$ is a belief state for the critic.

\paragraph{Model Architecture.}
We adapt the V\&L BERT architecture \citep{hong2020recurrent} for modeling the execution policy $\exepol$.
Our model has two components: an encoder and a decoder; both are implemented as Transformer models \citep{vaswani2017attention}.
The encoder takes as input a description $d^s_t$ or $d^g_t$ and generates a sequence of hidden vectors.
In every step, the decoder takes as input the previous hidden vector $b^s_{t - 1}$, the sequence of vectors representing $d^s_t$, and the sequence of vectors representing $d^g_t$.
It then performs self-attention on these vectors to compute the current hidden vector $b^s_{t}$ and a probability distribution over navigation actions $p_t$.

The \intpoltxt $\inpol$ (the actor) is an LSTM-based recurrent neural network.
The input of this model is the execution policy's model outputs, $b^s_{t}$ and $p_t$, and the embedding of the previously taken action $\b a_{t - 1}$.
The critic model also has a similar architecture but outputs a real number (the $V$ value) rather than an action distribution.
When training the \intpoltxt, we always fix the parameters of the execution policy.
We find it necessary to pre-train the critic before training it jointly with the actor. \looseness=-1

\paragraph{Representation of State Descriptions.}

The representation of each object, room, or action is computed as follows.
Let  $f^{\textrm{name}}$, $f^{\textrm{horz}}$, $f^{\textrm{vert}}$, $f^{\textrm{dist}}$, and $f^{\textrm{type}}$ are the features of an object $f$, consisting of its name, horizontal angle, vertical angle, distance, and type (a type is either \texttt{Object}, \texttt{Room}, or \texttt{Action}; in this case, the type is \texttt{Object}). 
For simplicity, we discretize real-valued features, resulting in 12 horizontal angles (corresponding to $\pi/6 \cdot k, 0 \leq k < 12$), 3 vertical angles (corresponding to $\pi/6 \cdot k, -1 \leq k \leq 1$), and 5 distance values (we round down a real-valued distance to the nearest integer). 
We then lookup the embedding of each feature from an embedding table and sum all the embeddings into a single vector that represents the corresponding object.
For a room, $f^{\textrm{horz}}$, $f^{\textrm{vert}}$ $f^{\textrm{dist}}$ are zeroes.
For an action, $f^{\textrm{name}}$ is either \texttt{ActionStop} for the stop action $a_{\textrm{done}}$ or \texttt{ActionGo} otherwise.

During the pre-training phase, we randomly drop features in $d^s_t$ and $d^g_t$ so that the execution policy is familiar with making decisions under sparse information.  
Concretely, we refer to all features of an object, room or action as a \textit{feature set}.
For $d^s_t$, let $M$ be the number of objects in a description. 
We uniformly randomly keep $m$ feature sets among the $M + 1$ feature sets of $d^s_t$ (the plus one is the room's feature set), where $m \sim \textrm{Uniform}(\min(5, M+1), M+1)$.

For $d^s_t$, we have two cases.
If $g_1$ is not adjacent or equals to $s_1$, we uniformly randomly alternate between giving a dense and a sparse description.
In this case, the sparse description contains the features of the target object and the goal room's name. 
Otherwise, with a probability of \sfrac{1}{3}, we give either  (a) a dense description (b) a (sparse) description that contains the target object's features and the goal room's name, or (c) a (sparse) description that describes the next ground-truth action.

We pre-train the execution policy on various path lengths (ranging from 1 to 10 graph nodes) so that it learns to accomplish both long-distance main goals and short-distance subgoals. 

\begin{table*}[t!]
\centering
\footnotesize
\caption{Dataset statistics.}
\vspace{-0.3cm}
\begin{tabular}{lcc}
    \toprule
    \multicolumn{1}{c}{Split} & \multicolumn{1}{c}{Number of examples} \\
    \midrule
    Pre-training & 82,104 \\
    Pre-training validation & 3,000 \\
    Training & 65,133 \\
    Validation \unseenstr & 1,901 \\
    Validation \unseenobj & 1,912 \\
    Validation \unseenenv & 1,967 \\
    Test \unseenstr & 1,653\\
    Test \unseenobj & 1,913 \\
    Test \unseenenv & 1,777 \\
    \bottomrule
\end{tabular}
\label{tab:data}
\end{table*}

\paragraph{Data.} \autoref{tab:data} summarizes the data splits. 
From a total of 72 environments provided by the Matterport3D dataset, we use 36 environments for pre-training, 18 as unseen environments for training, 7 for validation \unseenenv, and 11 for test \unseenenv.
We use a vocabulary of size 1738, which includes object and room names, and special tokens representing the distance and direction values. 
The length of a navigation path ranges from 5 to 10 graph nodes.

\begin{table*}[t!]
\centering
\footnotesize
\caption{Hyperparameters.}
\vspace{-0.3cm}
\begin{tabular}{lc}
    \toprule
    \multicolumn{1}{c}{Hyperparameter Name} & \multicolumn{1}{c}{Value} \\
    \midrule
    \multicolumn{2}{l}{\textbf{Environment}} \\
    Max. subgoal distance ($l_\textrm{max}$) & 3 nodes \\
    Max. stack size ($L$) & 2 \\
    Max. object distance for $d^s_t$ & 5 meters \\
    Max. object distance for $d^g_t$ & 3 meters \\
    Max. number of objects ($M_{\textrm{max}}$) & 20 \\
    Cost of taking each \cur, \goal, \sub, \doa action & 0.01 \\
    \midrule
    \multicolumn{2}{l}{\textbf{Execution policy $\exepol$}}  \\
    Hidden size & 256 \\
    Number of hidden layers & 2 \\
    Attention dropout probability & 0.1 \\
    Hidden dropout probability & 0.1 \\ 
    Number of attention heads & 8 \\ 
    Optimizer & Adam \\ 
    Learning rate & $10^{-4}$ \\
    Batch size & 32 \\
    Number of training iterations & $10^5$ \\
    Max. number of time steps ($H$) & 15 \\
    \midrule
    \multicolumn{2}{l}{\textbf{Intention policy $\inpol$}} \\
    Hidden size & 512 \\
    Number of hidden layers & 1 \\
    Entropy regularization weight & 0.001 \\
    Optimizer & Adam \\ 
    Learning rate & $10^{-5}$ \\
    Batch size & 32 \\
    Number of critic pre-training iterations & $5 \times 10^3$ \\
    Number of training iterations & $5\times10^4$ \\
    Max. number of time steps ($H$) & 30 \\
    Max. number of time steps for executing a subgoal & 3$\times$ shortest distance to the subgoal \\
    \bottomrule
\end{tabular}
\label{tab:hyper}
\end{table*}

\paragraph{Hyperparameters.} See \autoref{tab:hyper}.